\documentclass{article}
\usepackage{ijcai20}

\usepackage{times}
\usepackage{soul}
\usepackage{url}
\usepackage[hidelinks]{hyperref}
\usepackage[utf8]{inputenc}
\usepackage[small]{caption}
\usepackage{graphicx}
\usepackage{amsmath}
\usepackage{amsthm}
\usepackage{booktabs}
\usepackage{algorithm}
\usepackage{algorithmic}
\usepackage{booktabs}
\usepackage{multirow}
\usepackage{threeparttable}
\usepackage{microtype}

\newcommand\citet[1]{\citeauthor{#1} [\citeyear{#1}]}
\usepackage{mymathdef}
\usepackage{subfig}
\usepackage{float}

\title{Improving BERT Fine-Tuning via Self-Ensemble and Self-Distillation}

\author{
Yige Xu$^1$
\and
Xipeng Qiu$^1$\and
Ligao Zhou$^2$\And
Xuanjing Huang$^1$
\affiliations
$^1$Shanghai Key Laboratory of Intelligent Information Processing, Fudan University\\
School of Computer Science, Fudan University\\
825 Zhangheng Road, Shanghai, China\\
$^2$Huawei Technologies Co., Ltd.
\emails
\{ygxu18, xpqiu, xjhuang\}@fudan.edu.cn,
zhouligao@huawei.com,
}

\begin{document}

\maketitle

\begin{abstract}
Fine-tuning pre-trained language models like BERT has become an effective way in NLP and yields state-of-the-art results on many downstream tasks.
Recent studies on adapting BERT to new tasks mainly focus on modifying the model structure, re-designing the pre-train tasks, and leveraging external data and knowledge.
The fine-tuning strategy itself has yet to be fully explored.
In this paper, we improve the fine-tuning of BERT with two effective mechanisms: \textit{self-ensemble} and \textit{self-distillation}. The experiments on text classification and natural language inference tasks show our proposed methods can significantly improve the adaption of BERT without any external data or knowledge.


\end{abstract}

\section{Introduction}
The pre-trained language models including
BERT~\cite{devlin2018bert} and its variants (XLNet~\cite{NIPS2019_8812} and RoBERTa \cite{liu2019roberta}) have been proven beneficial for many natural language processing (NLP) tasks, such as text classification, question answering~\cite{rajpurkar2016squad} and natural language inference~\cite{snli:emnlp2015}. These pre-trained models have learned general-purpose language representations on a large amount of unlabeled data, therefore, adapting these models to the downstream tasks can bring a good initialization for and avoid training from scratch.  There are two common ways to utilize these pre-trained models on downstream tasks:
feature extraction (where the pre-trained parameters are frozen), and fine-tuning (where the pre-trained parameters are unfrozen and fine-tuned). Although both these two ways can significantly improve the performance of most of the downstream tasks, the fine-tuning way usually achieves better results than feature extraction way~\cite{peters2019tune}.
Therefore, it is worth paying attention to find a good fine-tuning strategy.

As a widely-studied pre-trained language model, the potential of BERT can be further boosted by
modifying model structure \cite{stickland2019bert,houlsby2019parameter}, re-designing pre-training objectives \cite{dong2019unified,NIPS2019_8812,liu2019roberta}, data augmentation \cite{2019t5} and optimizing fine-tuning strategies with external knowledge \cite{liu2019multi,sun2019fine}.
However, the fine-tuning strategy itself has yet to be fully explored. \citet{sun2019fine} investigated different fine-tuning strategies and hyper-parameters of BERT for text classification and showed the ``further-pretraining'' on the related-domain corpus can further improve the ability of BERT. \citet{liu2019multi} fine-tuned BERT under the multi-task learning framework. The performance of BERT on a task could benefit from other related tasks. Although these methods achieve better performance, they usually need to leverage external data or knowledge.

In this paper, we investigate how to maximize the utilization of BERT by better fine-tuning strategy without utilizing the external data or knowledge. BERT is usually fine-tuned by using stochastic gradient descent (SGD) method. In practice, the performance of fine-tuning BERT is often sensitive to the different random seeds and orders of training data, especially when the last training sample is noise.
To alleviate this, the ensemble method is widely-used to combine several fine-tuned based models since it can reduce the overfitting and improve the model generalization. The ensemble BERT
usually achieves superior performance than the single BERT model.
However, the main disadvantages of the ensemble method are its model size and training cost. The ensemble model needs to keep multiple fine-tuned BERTs and has a low computation efficiency and high storage cost.

 We improve the fine-tuning strategy of BERT by introducing two mechanisms: \textit{self-ensemble} and \textit{self-distillation}.

(1) \textbf{Self-Ensemble}. Motivated the success of widely-used ensemble models, we propose a self-ensemble method, in which the base models are the intermediate BERT models at different time steps within a single training process~\cite{polyak1992acceleration}. To further reduce the model complexity of the ensemble model, we use a more efficient ensemble method, which combines several base models with parameter averaging rather than keeping several base models.

(2) \textbf{Self-Distillation}. Although the self-ensemble can improve the model performance, the training of the base model is the same as the vanilla fine-tuning strategy and cannot be affected by the ensemble model. We further use knowledge distillation~\cite{hinton2015distilling} to improve fine-tuning efficiency. At each time step in training, the current BERT model (called \textit{student model}) is learned with two teachers: the gold labels and self-ensemble model (called \textit{teacher model}). The teacher model is an average of student models at previous time steps. With the help of the teacher model, the student is more robust and accurate. Moreover, a better student model further leads to a better teacher model. A similar idea is also used in semi-supervised learning, such as Temporal Ensembling~\cite{laine2016temporal} and Mean Teacher ~\cite{tarvainen2017mean}. Different from them, our proposed self-distillation aims to optimize the student model without external data.

The experiments on text classification and natural language inference tasks show our proposed methods can reduce the test error rate by more than $5.5\%$ on the average on seven widely-studied datasets.

The contributions of this paper are summarized as follows:
 \begin{itemize*}
   \item We show the potential of BERT can be further stimulated by a better fine-tuning strategy without leveraging external knowledge or data.
   \item The self-ensemble method with parameter averaging can improve BERT without significantly decreasing the training efficiency.
   \item With self-distillation, the student and teacher models can benefit from each other. The distillation loss can also be regarded as a regularization to improve the generalization ability of the model.
 \end{itemize*}

\section{Related Work}
We briefly review two kinds of related work: pre-trained language models and knowledge distillation.

\subsection{Pre-trained Language Models}
\label{section-bert}

Pre-training language models on a large amount of unlabeled data then fine-tuning in downstream tasks has become a new paradigm for NLP and made a breakthrough in many NLP tasks. Most of the recent pre-trained language models (e.g., BERT \cite{devlin2018bert}, XLNet \cite{NIPS2019_8812} and RoBERTa \cite{liu2019roberta}) are built with Transformer architecture~\cite{vaswani2017attention}.

As a wide-used model, BERT is pre-trained on \textit{Masked Language Model Task} and \textit{Next Sentence Prediction Task} via a large cross-domain unlabeled corpus. BERT has two different model size: $\mathrm{BERT}_{\mathrm{BASE}}$ with a 12-layer Transformer encoder and $\mathrm{BERT}_{\mathrm{LARGE}}$ with a 24-layer Transformer encoder. Both of them take an input of a sequence of no more than 512 tokens and outputs the representation of the sequence. The sequence has one segment for text classification task or two for text matching task. A special token \texttt{[CLS]} is added before segments, which contain the special classification embedding. Another special token \texttt{[SEP]} is used for separating segments.

\paragraph{Fine-tuning}
BERT can deal with different natural language tasks with task-specific output layers. For text classification or text matching, BERT takes the final hidden state $\bh$ of the first token \texttt{[CLS]} as the representation of the input sentence or sentence-pair. A simple softmax classifier is added to the top of BERT to predict the probability of label $y$:
\begin{align}
p(y|\bh) =\mathrm{softmax}(W\bh),
\end{align}
where $W$ is the task-specific parameter matrix. The cross-entropy loss is used to fine-tune BERT as well as $W$ jointly.

\subsection{Knowledge Distillation for Pre-trained Models}

Since the pre-trained language models usually have an extremely large number of parameters, they are difficult to be deployed on the resource-restricted devices. Several previous works leverage the knowledge distillation~\cite{hinton2015distilling}
approach to reducing model size while maintaining accuracy, such as TinyBERT~\cite{jiao2019tinybert} and DistilBERT
~\cite{sanh2019distilbert}.
Knowledge Distillation aims to transfer the
knowledge of a large teacher model to a small student model by training the student model to reproduce the behaviors of the teacher model.

The teacher model usually is well-trained and fixed in the processing knowledge distillation. Unlike the common way of knowledge distillation, we perform knowledge distillation in online fashion. The teacher model is an ensemble of several student models at previous time steps within the fine-tuning stage.



\section{Methodology}

The fine-tuning of BERT usually aims to minimize the cross-entropy loss on a specific task with stochastic gradient descent method. Due to the stochastic nature, the performance of fine-tuning is often affected by the random orderings of training data, especially when the last training samples is noise.

Our proposed fine-tuning strategy is motivated by the ensemble method and knowledge distillation. There are two models in our fine-tuning strategy: a student model is the fine-tuning BERT and a teacher model is a self-ensemble of several student models. At each time step, we further distillate the knowledge of the teacher model to the student model.

\subsection{Ensemble BERT}\label{section-ensemble-bert}

In practice, the ensemble method is usually adopted to further improve the performance of BERT.

\paragraph{Voted BERT}
The common ensemble method is voting-based.
We first fine-tune multiple BERT with different random seeds. For each input, we output the best predictions made by the fine-tuned BERT along
with the probability and sum up the probability of predictions from each model together. The output of the ensemble model is the prediction with the highest
probability.

Let $\mathrm{BERT}(x;\theta_k)\ (1\leq k\leq K)$ are $K$ BERT models fine-tuned on a specific task with different random seeds, the ensemble model $\mathrm{BERT}_{\mathrm{VOTE}}(x;\Theta)$ is defined as
\begin{align}
\mathrm{BERT}_{\mathrm{VOTE}}(x;\Theta)=\sum_{k=1}^{K} \mathrm{BERT}(x;\theta_k),
\end{align}
where $\theta_k$ denotes the parameters of the $k$-th model and $\Theta$ denotes the parameters of all the $K$ models.

We call this kind of ensemble model as \textit{voted BERT}. The voted BERT can greatly boost the performance of single BERT in many tasks, such as question answering~\cite{rajpurkar2016squad}. However, a major disadvantage of the voted BERT is it needs keeping multiple different BERTs and its efficiency is low in computing and storage.

\paragraph{Averaged BERT}

To reduce the model complexity of ensemble model, we use a parameter-averaging strategy to combine several BERTs into a single model, called \textit{averaged BERT}. The averaged BERT is defined as
\begin{align}
\mathrm{BERT}_{\mathrm{AVG}}(x;\bar{\theta})= \mathrm{BERT}(x;\frac{1}{K}\sum_{k=1}^{K} \theta_k),
\end{align}
where $\bar{\theta}$ is the averaged parameters of $K$ individual fine-tuned BERTs.

\begin{figure}[t]
\centering
\subfloat[Voted BERT]{
\includegraphics[width=0.6\linewidth]{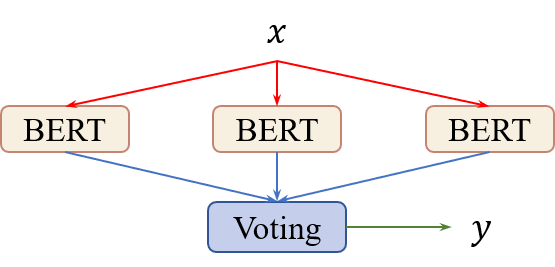}
}\\
\subfloat[Averaged BERT]{
\includegraphics[width=0.6\linewidth]{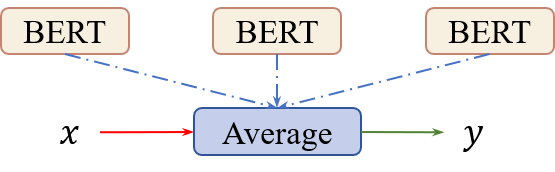}
}
	\caption{Two methods to ensemble BERT. The averaged BERT has better computational and memory efficiency then voted BERT.}\label{fig:ensemble}
\end{figure}

Figure \ref{fig:ensemble} illustrates two kinds of ensemble BERTs. Since the averaged BERT is indeed a single model and has better computational and memory efficiency than the voted BERT.

\subsection{Self-Ensemble BERT}\label{section-self-avg-bert}

Although the ensemble BERT usually brings better performance, it needs to train multiple BERTs and its cost is often expensive. To reduce the training cost, we use a self-ensemble model to combine the intermediate models at different time steps in a single training phase. We regard the BERT at each time step as a base model and combine them into a self-ensemble model.

Here we only describe the self-ensemble BERT with parameter averaging, since the voted version of self-ensemble is impracticable to keep all the intermediate models.

Let $\theta_t$ denote parameters when fine-tuning BERT at time step $t$, the self-ensemble BERT is defined as
  \begin{align}\label{eqation-BERT-self-average}
    \mathrm{BERT}_{\mathrm{SE}}(x;\bar{\theta})={\mathrm{BERT}(x;\frac{1}{T}\sum_{\tau=1}^{T}\theta_t)},
  \end{align}
where $\bar{\theta}$ is the averaged parameters of BERTs over $T$ time steps.

Averaging model weights over training steps tends to produce a more accurate model than using the final weights directly \cite{polyak1992acceleration}.

\begin{figure}[t!]
\centering
\includegraphics[width=\linewidth]{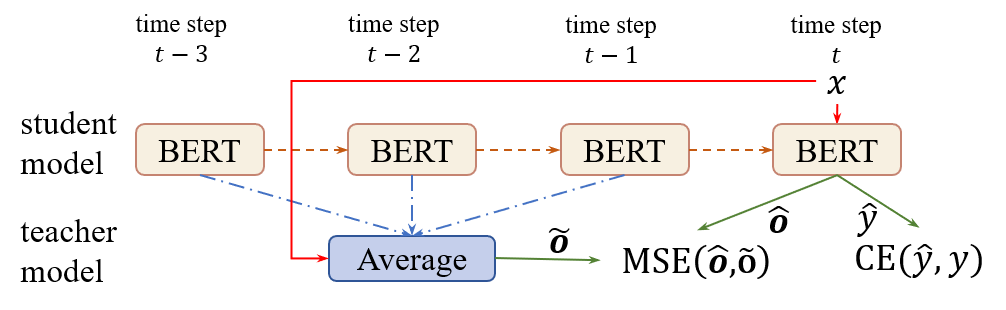}
	\caption{The proposed fine-tuning strategy $\mathrm{BERT}_{\mathrm{SDA}}$. The teacher model is the average of student models over recent $K$ time steps, and $K=3$ here. $(x,y)$ is the input training sample. $\hat{y}$ is the probability of label $y$ predicted by student model. $\hat{\mathbf{o}}$ and $\tilde{\mathbf{o}}$ are the logits output by student model and teacher model reflectively. MSE and CE denote the mean square error and cross-entropy loss.}\label{fig:proposed}
\end{figure}

\subsection{Self-Distillation BERT}\label{section-self-distillation-bert}

Although the self-ensemble can improve the model performance, the base model is trained in the same manner to the vanilla fine-tuning strategy and cannot be affected by the ensemble model. We further use knowledge distillation~\cite{hinton2015distilling} to improve the base model. At each time step in training, the current BERT model (called \textit{student model}) is learned from two teachers: the gold labels and the self-ensemble model (called \textit{teacher model}). The teacher model is an average of student models at previous time steps. With the help of the teacher model, the student is more robust and accurate.

\paragraph{Self-Distillation-Averaged (SDA)}

We first denote a fine-tuning strategy $\mathrm{BERT}_{\mathrm{SDA}}$, in which the teacher model is self-ensemble BERT with parameter averaging.

Let $\mathrm{BERT}(x,\theta)$ denote the student BERT, the objective of $\mathrm{BERT}_{\mathrm{SDA}}$ strategy is
\begin{align}
\mathcal{L}_{\theta}(x, y)=&\mathrm{CE}\Big(\mathrm{BERT}(x,\theta),y \Big)\nonumber\\
+&\lambda\mathrm{MSE}\Big(\mathrm{BERT}(x,\theta),\mathrm{BERT}(x,\bar{\theta})\Big),
\end{align}
where $\mathrm{CE}$ and $\mathrm{MSE}$ denote the cross-entropy loss and mean squared error respectively, and $\lambda$ balances the importance of two loss functions.  The teacher model $\mathrm{BERT}(x,\bar{\theta})$ is a self-ensemble BERT with recent time steps. At time step $t$,  $\bar{\theta}$ is the averaged parameters of recent $K$ time steps:
\begin{align}\label{equation-sda-params}
\bar{\theta} = \frac{1}{K}\sum_{k=1}^{K} \theta_{t-k},
\end{align}
where $K$ is a hyperparameter denoting the teacher size.

Figure \ref{fig:proposed} shows the training phase of our proposed method.
In the training phase, we can compute $\bar{\theta}$ efficiently by moving average.

Since the teacher model aggregates information of student models after every
time step, it is usually more robust. Moreover, a better student model further leads to better a teacher model.

\paragraph{Self-Distillation-Voted (SDV)}

As a comparison, we also propose an alternative self-distillation method by replacing the teacher model with self-voted BERT, called $\mathrm{BERT}_{\mathrm{SDV}}$. The objective of $\mathrm{BERT}_{\mathrm{SDV}}$ strategy is
\begin{align}
&\mathcal{L}_{\theta}(x, y)=\mathrm{CE}\Big(\mathrm{BERT}(x,\theta),y \Big)\nonumber\\
& +\lambda\mathrm{MSE}\Big(\mathrm{BERT}(x,\theta),\frac{1}{K}\sum_{k=1}^{K} \mathrm{BERT}(x,\theta_{t-k})\Big).\label{equation-sdv-loss}
\end{align}

\begin{table*}[t!]\small
    \centering
    \begin{tabular}{c| c c c c c c}
    \toprule
    Type & Dataset & Num of Labels & Train samples & Dev samples & Test samples \\
    \midrule
    \multirow{5} * {Text Classification} & IMDb & 2 & 25,000 & - & 25,000  \\
    ~ & AG's News & 4 & 120,000 & - & 7,600 \\
    ~ & DBPedia & 14 & 560,000 & - & 70,000 \\
    ~ & Yelp Polarity & 2 & 560,000 & - & 38,000 \\
    ~ & Yelp Full & 5 & 650,000 & - & 50,000 \\
    \midrule
    \multirow{2} * {NLI} & SNLI & 3 & 549,367 & 9,842 & 9,824  \\
    ~ & MNLI-(m/mm) & 3 & 392,702 & 9,815/9,832 & 9,796/9,847 \\
    \bottomrule
\end{tabular}
		\caption{\label{table_stat} Summary statistics of seven widely-studied text classification and natural language inference (NLI) datasets.
		}
	\end{table*}

The training efficiency of $\mathrm{BERT}_{\mathrm{SDV}}$ strategy is lower than $\mathrm{BERT}_{\mathrm{SDA}}$ strategy since $\mathrm{BERT}_{\mathrm{SDV}}$ needs to process the input with recent $K$ student models.

\section{Experiments}

In this paper, we improve BERT fine-tuning via self-ensemble and self-distillation. The vanilla fine-tuning method of BERT is used as our baseline. Then we evaluate our proposed fine-tuning strategies on seven datasets to demonstrate the feasibility of our self-distillation model.

\subsection{Datasets}
 Our proposed method is evaluated on five Text Classification datasets and two Natural Language Inference (NLI) datasets. The statistics of datasets are shown in Table \ref{table_stat}.

\paragraph{Text Classification}
\begin{itemize*}
  \item \textbf{IMDb}
  IMDb \cite{maas2011learning} is a binary sentiment analysis dataset from the Internet Movie Database. The dataset has 25,000 training examples and 25,000 validation examples. The task is to predict whether the review text is positive or negative.

  \item \textbf{AG's News}
  AG's corpus \cite{zhang2015character} of the news articles on the web contains 496,835 categorized news articles. The four largest classes from this corpus with only the title and description fields were chosen to construct the AG's News dataset.

  \item \textbf{DBPedia}
  DBPedia \cite{zhang2015character} is a crowd-sourced community effort that includes structured information from Wikipedia. The DBPedia dataset is constructed by picking 14 non-overlapping classes from DBPedia 2014 with only the title and abstract of each Wikipedia article.

  \item \textbf{Yelp}
  The Yelp dataset is obtained from the Yelp Dataset Challenge in 2015, built by \citet{zhang2015character}. There are two classification tasks in this dataset: Yelp Full and Yelp Polarity. Yelp Full predicts the full number of stars (1 to 5) which given by users, and the other predicts a polarity label that is positive or negative.

\end{itemize*}

\paragraph{Natural Language Inference}
\begin{itemize*}
  \item \textbf{SNLI}
  The Stanford Natural Language Inference Corpus \cite{snli:emnlp2015} is a collection of 570k human-written English sentence pairs manually labeled for balanced classification with the labels entailment, contradiction, or neutral.

	\item \textbf{MNLI}
	Multi-Genre Natural Language Inference \cite{N18-1101} corpus is a crowdsourced entailment classification task with about 433k sentence pairs. 
The corpus has two test sets: matched (MNLI-m) and mismatched (MNLI-mm).

\end{itemize*}

\subsection{Hyperparameters}

All the hyperparameters of our proposed methods are the same as the official BERT~\cite{devlin2018bert} except self-distillation weight $\lambda$ and teacher size $K$. We use AdamW optimizer with the warm-up proportion of 0.1, base learning rate for BERT encoder of 2e-5, base learning rate for softmax layer of 1e-3, dropout probability of 0.1. For sequences of more than 512 tokens, we truncate them and choose head 512 as model input.

We fine-tune all models on one RTX 2080Ti GPU.
For $\mathrm{BERT}_{\mathrm{BASE}}$, the batch size is 4, and the gradient accumulation steps is 4.
For $\mathrm{BERT}_{\mathrm{LARGE}}$, the batch size is 1, and the gradient accumulation steps is 16.

For ensemble BERT (See section \ref{section-ensemble-bert}), we run $\mathrm{BERT}_{\mathrm{BASE}}$ with 4 different random seeds and save the checkpoint.

\subsection{Model Selection}\label{experiment-hyper}

As shown in section \ref{section-self-distillation-bert}, there are  two main hyperparameters in our fine-tuning methods ($\mathrm{BERT}_{\mathrm{SDA}}$ and $\mathrm{BERT}_{\mathrm{SDV}}$): self-distillation weight $\lambda$ and teacher size $K$.

\paragraph{Self-Distillation Weight}\label{experiment-different-sd-norm}

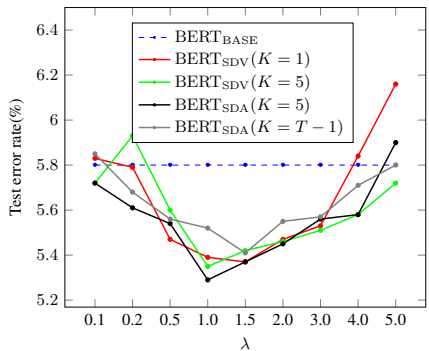
\begin{figure}[h!]
    \centering
    \begin{tikzpicture}[scale=0.7]
    \pgfplotstableread{./res/distillation_norm.txt}\datatable
      \begin{axis}[
      font=\small,
      xlabel={$\lambda$},
      xtick=data,
        xticklabels from table={\datatable}{[index] 0},
        xtick distance=1,
        table/x expr=\coordindex,
      ymax=6.5,
      ylabel={Test error rate(\%)},
      legend entries={$\mathrm{BERT}_{\mathrm{BASE}}$,
      $\mathrm{BERT}_{\mathrm{SDV}}(K=1)$, $\mathrm{BERT}_{\mathrm{SDV}}(K=5)$, $\mathrm{BERT}_{\mathrm{SDA}}(K=5)$,
      $\mathrm{BERT}_{\mathrm{SDA}}(K=T-1)$
      },
      mark size=1.0pt,
      legend pos= north east,
      legend cell align={left}, 
      legend style={font=\small,line width=.5pt,mark size=.5pt,
              at={(0.8,0.95)},
              /tikz/every even column/.append style={column sep=0.5em}},
      ],

    \addplot [blue,dashed,mark=*] table [x=data, y=BERT-O] from \datatable;

      \addplot [red,thick,mark=*] table [x=data, y=BERT-SDV-1] from \datatable;
			\addplot [green,thick,mark=*] table [x=data, y=BERT-SDV-5] from \datatable;
			\addplot [black,thick,mark=*] table [x=data, y=BERT-SDA-5] from \datatable;
		\addplot [gray,thick,mark=*] table [x=data, y=BERT-SDA-0] from \datatable;
      \end{axis}
  \end{tikzpicture}
    \caption{Test error on IMDb dataset over different setting of self-distillation weight $\lambda$. $T$ denotes the total number of iterations.}\label{fig:exp-sd-norm}
  \end{figure}

We first evaluate our methods on IMDb dataset to investigate the effect of self-distillation weight $\lambda$. Figure \ref{fig:exp-sd-norm} shows that $ \lambda \in [1.0,1.5]$ has better results. This observation is the same as the other datasets. Therefore, we set $\lambda=1$ in the following experiments.

\paragraph{Teacher Size}
\label{experiment-teacher-size}

We choose different teacher size $K$ and evaluate our models in three datasets. Table \ref{table-exp-hyper-k} shows that teacher size is sensitive to datasets.
Therefore, we select the best teacher size for each dataset in the following experiment.

\begin{table}[h!]\small
	\centering
	\begin{tabular}{l |c| c c c}
	\toprule
	\multirow{2}*{Model} & \multirow{2}*{$K$} & \multirow{2}*{IMDb} & AG's & \multirow{2}*{SNLI}\\
	~ & ~ & ~ & News & ~ \\
	\midrule
	\multirow{4} * {$\mathrm{BERT}_{\mathrm{SDV}}$} & 1 & 5.39 & \textbf{5.38} & \textbf{91.2} \\
	~ & 2 & 5.44 & 5.39 & 91.1 \\
	~ & 3 & 5.40 & 5.50 & \textbf{91.2} \\
	~ & 4 & 5.47 & 5.49 & \textbf{91.2} \\
	~ & 5 & \textbf{5.35} & 5.55 & 91.1 \\
	\midrule
	\multirow{4} * {$\mathrm{BERT}_{\mathrm{SDA}}$} & $T-1$ & 5.41 & \textbf{5.29} & 91.0 \\
	~ & 2 & 5.46 & 5.49 & \textbf{91.2} \\
	~ & 3 & 5.48 & 5.55 & 91.1 \\
	~ & 4 & 5.44 & 5.52 & 91.1 \\
	~ & 5 & \textbf{5.29} & 5.41 & 91.1 \\
	\bottomrule
\end{tabular}
\caption{Results on IMDb dataset over different teacher sizes. $\mathrm{BERT}_{\mathrm{SDV}}(K=1)$ is same as  $\mathrm{BERT}_{\mathrm{SDA}}(K=1)$. For IMDb and AG's News, we report test error rate (\%). For SNLI, we report accuracy (\%). $T$ denotes the total number of iterations.}
\label{table-exp-hyper-k}
\end{table}

\subsection{Model Analysis}\label{experiment-sd-epoch}
\paragraph{Training Stability}\label{experiment-training-stability}

Generally, distinct random seeds can lead to substantially different results when fine-tuning BERT even with the same hyperparameters.  Thus, we conduct experiments to explore the effect of data order on our models.

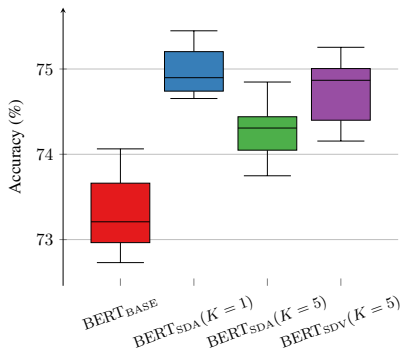
\begin{figure}[ht!]
\centering
\begin{tikzpicture}[scale=0.65]
	\pgfplotstableread[col sep=comma]{./res/boxplot.txt}\csvdata
	\pgfplotstabletranspose\datatransposed{\csvdata}
	\begin{axis}[
		boxplot/draw direction = y,
		x axis line style = {opacity=0},
		axis x line* = bottom,
		axis y line = left,
		enlarge y limits,
		ymajorgrids,
		xtick = {1, 2, 3, 4},
		xticklabel style = {align=center, font=\small, rotate=20},
		xticklabels = {$\mathrm{BERT}_{\mathrm{BASE}}$, $\mathrm{BERT}_{\mathrm{SDA}}(K=1)$, $\mathrm{BERT}_{\mathrm{SDA}}(K=5)$, $\mathrm{BERT}_{\mathrm{SDV}}(K=5)$},
		ylabel = {Accuracy (\%)},
		ytick = {73, 74, 75}
	]
		\foreach \n in {1,...,4} {
			\addplot+[boxplot, fill, draw=black] table[y index=\n] {\datatransposed};
		}
	\end{axis}
\end{tikzpicture}

	\caption{Boxplot for training stability. Only 1,500 labeled example in SNLI training set are used for training. Here $\mathrm{BERT}_{\mathrm{SDA}}(K=1)$ is same as $\mathrm{BERT}_{\mathrm{SDV}}(K=1)$. }
\label{fig:exp-small-snli}
\end{figure}

This experiment is conducted with a set of data order seeds. One data order can be regarded as one sample from the set of permutations of the training data. 1,500 labeled examples from SNLI dataset (500 on each class) are randomly selected to construct a new training set. With the same initialization seeds but different data order seeds, we run 10 times for each fine-tuning strategy and record the result as Figure~\ref{fig:exp-small-snli}.

Results show that our strategies have higher accuracy and smaller variance than the vanilla $\mathrm{BERT}_{\mathrm{BASE}}$ fine-tuning. This proves that the fine-tuned BERT with the self-distillation strategy inherits the property of the ensemble model and is less sensitive to the data order.

\paragraph{Convergence Curves}
\label{experiment-training-curves}

\begin{figure}[ht!]
    \centering
    \begin{tikzpicture}[scale=0.65]
    \pgfplotstableread{./res/training_curves.txt}\datatable
      \begin{axis}[
      font=\small,
      xlabel={Epoch},
      xtick=data,
        xticklabels from table={\datatable}{[index] 0},
        xtick distance=1,
        table/x expr=\coordindex,
      ylabel={Test error rate(\%)},
      legend entries={$\mathrm{BERT}_{\mathrm{BASE}}$,
      $\mathrm{BERT}_{\mathrm{SDV}}(K=5)$,
      $\mathrm{BERT}_{\mathrm{SDA}}(K=5)$,
      $\mathrm{BERT}_{\mathrm{SDA}}(K=T-1)$},
      mark size=1.0pt,
      ymajorgrids=true,
      grid style=dashed,
      legend pos= north east,
      legend cell align={left}, 
      legend style={font=\small,line width=.5pt,mark size=.5pt,
              /tikz/every even column/.append style={column sep=0.5em}},
              smooth,
      ],

      \addplot [red,thick,mark=*] table [x=data, y=BERT-original] from \datatable;
      \addplot [black,thick,mark=*] table [x=data, y=BERT-SDV-5] from \datatable;
				\addplot [green,thick,mark=*] table [x=data, y=BERT-SDA-5] from \datatable;
        \addplot [blue,thick,mark=*] table [x=data, y=BERT-SDA-0] from \datatable;


      \end{axis}
  \end{tikzpicture}
  \caption{Test error rates(\%) on IMDb dataset durning different epoch. $T$ denotes the total number of iterations.}\label{fig:exp-training-curves}
  \end{figure}
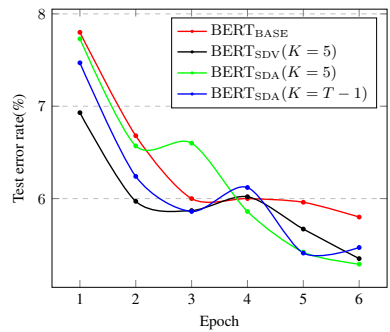

	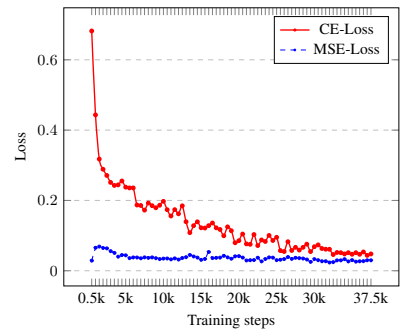
\begin{figure}[ht!]
	    \centering
	    \begin{tikzpicture}[scale=0.65]	    \pgfplotstableread{./res/distillation_loss.txt}\datatable
	      \begin{axis}[
	      font=\small,
	      xlabel={Training steps},
	      xtick=data,
	      xticklabels={0.5k,,,,,,,,,5k,,,,,,,,,,10k,,,,,,,,,,15k,,,,,,,,,,20k,,,,,,,,,,25k,,,,,,,,,,30k,,,,,,,,,,,,,,,37.5k},
	        xtick distance=1,
	        table/x expr=\coordindex,
	      ylabel={Loss},
	      legend entries={CE-Loss, MSE-Loss},
	      mark size=1.0pt,
	      ymajorgrids=true,
	      grid style=dashed,
	      legend pos= north east,
	      legend style={font=\small,line width=.5pt,mark size=.5pt,
	              /tikz/every even column/.append style={column sep=0.5em}},
	              smooth,
	      ]

	      \addplot [red,thick,mark=*] table [x=data, y=CE-LOSS] from \datatable;
				\addplot [blue,dashed,mark=*] table [x=data, y=MSE-LOSS] from \datatable;


	      \end{axis}
	  \end{tikzpicture}
	    \caption{Loss curve of $\mathrm{BERT}_{\mathrm{SDA}}(K=1)$ on IMDb dataset.}\label{fig:exp-sd-loss}
	  \end{figure}

\begin{table*}[t!]\small
\setlength{\tabcolsep}{4pt}
    \centering
    \begin{tabular}{l |c c c c c| c|| c c| c}
    \toprule
    \multirow{3}*{Model} & \multirow{2}*{IMDb} & AG's & \multirow{2}*{DBPedia} & \multirow{2}*{Yelp P.} & \multirow{2}*{Yelp F.} & \multirow{3}*{Avg. $\Delta$} & \multirow{2}*{SNLI} & MNLI & \multirow{3}*{Avg. $\Delta$}\\
		~ & ~ & News & ~ & ~ & ~ & ~ & ~ & (m/mm) & ~ \\
		\cline{2-6}\cline{8-9}
		~ & \multicolumn{5}{c|}{Test Error Rate (\%)} & ~ & \multicolumn{2}{c|}{Accuracy (\%)} & ~ \\
    \midrule
		ULMFiT \cite{howard2018universal} & 4.60 & 5.01 & 0.80 & 2.16 & 29.98 & / & / & / & / \\
		$\mathrm{BERT}_{\mathrm{BASE}}$ \cite{sun2019fine}* & 5.40 & 5.25 & 0.71 & 2.28 & 30.06 & / & / & / & / \\
		\midrule
		\midrule
    $\mathrm{BERT}_{\mathrm{BASE}}$ & 5.80 & 5.71 & 0.71 & 2.25 & 30.37 & - & 90.7 & 84.6/83.3 & - \\
    $\mathrm{BERT}_{\mathrm{VOTE}}$ ($K=4$) & 5.60 & 5.41 & 0.67 & 2.03 & 29.44 & 5.44\% & 91.2 & 85.3/84.4 & 5.50\% \\
		$\mathrm{BERT}_{\mathrm{AVG}}$ ($K=4$) & 5.68 & 5.53 & 0.68 & 2.03 & 30.03 & 4.07\% & 90.8 & 85.1/84.2 & 3.24\% \\
		\midrule
    \midrule
		$\mathrm{BERT}_{\mathrm{SE}}$ (ours) & 5.82 & 5.59 & 0.65 & 2.19 & 30.48 & 2.50\% & 90.8 & 84.2/83.3 & -0.51\% \\
		\midrule
		$\mathrm{BERT}_{\mathrm{SDV}}$ (ours)& 5.35 & 5.38 & \textbf{0.68} & 2.05 & \textbf{29.88} & 5.65\% & \textbf{91.2} & \textbf{85.3/84.3} & \textbf{5.30\%} \\
		$\mathrm{BERT}_{\mathrm{SDA}}$ (ours) & \textbf{5.29} & \textbf{5.29} & \textbf{0.68} & \textbf{2.04} & \textbf{29.88} & \textbf{6.26\%} & 91.2 & 85.0/84.3 & 4.65\% \\
    \bottomrule
\end{tabular}
		\caption{\label{table_exp-1} Effects on fine-tuning the BERT-base model ($\mathrm{BERT}_{\mathrm{BASE}}$). `*' indicates using extra fine-tuning strategies and data preprocessing.
`/' means no available reported result.
We implemented a ``$\mathrm{BERT}_{\mathrm{BASE}}$'' without any extra fine-tuning strategy as our baseline. ``$\mathrm{BERT}_{\mathrm{VOTE}}$'' and ``$\mathrm{BERT}_{\mathrm{AVG}}$'' means ensemble BERT (See section \ref{section-ensemble-bert}).
		``$\mathrm{BERT}_{\mathrm{SE}}$'' means self-ensemble BERT (See section \ref{section-self-avg-bert}). ``$\mathrm{BERT}_{\mathrm{SDV}}$'' and ``$\mathrm{BERT}_{\mathrm{SDA}}$'' means self-distillation BERT (See section \ref{section-self-distillation-bert}).
`Avg. $\Delta$' means the average of relative change, respectively.
We \textbf{bold} the better self-distillation results.
		}
	\end{table*}

To understand the effects of using self-distillation, we record the converge curve while training. The training curves on IMDb dataset are shown in Figure \ref{fig:exp-training-curves}. Fine-tuning $\mathrm{BERT}_{\mathrm{BASE}}$ cannot get significant improvement in the last 3 epochs (from $6.00\%$ to $5.80\%$).  But with self-distillation mechanisms, the test error rate can further decrease to $5.35\%$ (for $\mathrm{BERT}_{\mathrm{SDV}}$) and $5.29\%$ (for $\mathrm{BERT}_{\mathrm{SDA}}$).

To further analyze the reason for this observation, we also record the loss curve of cross-entropy (CE) loss and mean-squared-error (MSE) loss, as shown in Figure \ref{fig:exp-sd-loss}. When training begins, the CE loss dominates the optimization objective.
In the last phase of training, the cross-entropy loss becomes small, and a large proportion of gain also comes from self-distillation. Therefore, although optimizing the CE loss at the end of the training phase cannot continue improving the performance of BERT, self-distillation with ensemble BERT as the teacher will continuously enhance the generalization and robustness of BERT.

\subsection{Model Performance}

In this section, we evaluate our proposed fine-tuning strategies for the BERT-base and BERT-large models on text classification and NLI tasks.

\paragraph{Effects on Fine-tuning BERT-Base}\label{experiment-main}

Table \ref{table_exp-1} shows the results of fine-tuning the BERT-base model on five text classification datasets and two NLI datasets. For ensemble BERT, both the voted BERT ($\mathrm{BERT}_{\mathrm{VOTE}}$) and averaged BERT ($\mathrm{BERT}_{\mathrm{AVG}}$) outperform the single BERT ($\mathrm{BERT}_{\mathrm{BASE}}$).  The average improvement of $\mathrm{BERT}_{\mathrm{VOTE}}$ is $5.44\%$ (for text classification) and $5.50\%$ (for NLI), while $\mathrm{BERT}_{\mathrm{AVG}}$ follows closely with $4.07\%$ and $3.24\%$.  $\mathrm{BERT}_{\mathrm{VOTE}}$ outperforms $\mathrm{BERT}_{\mathrm{AVG}}$ on all tasks, which adheres to our intuition since $\mathrm{BERT}_{\mathrm{VOTE}}$ is more complicated.

The self-ensemble BERT ($\mathrm{BERT}_{\mathrm{SE}}$) has a slight improvement in classification tasks of $2.50\%$, but it does not work on NLI tasks. This is also a reason why we need self-distillation to improve the base models.

Overall, self-distillation model has significant improvement on both classification and NLI tasks. Table \ref{table_exp-1} shows that $\mathrm{BERT}_{\mathrm{SDA}}$ and $\mathrm{BERT}_{\mathrm{SDV}}$ outperform $\mathrm{BERT}_{\mathrm{BASE}}$ on all datasets.
Generally speaking, $\mathrm{BERT}_{\mathrm{SDA}}$ performs better than $\mathrm{BERT}_{\mathrm{SDV}}$ on text classification tasks with the improvement of $6.26\%$ vs. $5.65\%$, but the latter performs better on NLI tasks ($\mathrm{BERT}_{\mathrm{SDA}}$ vs. $\mathrm{BERT}_{\mathrm{SDV}}$ is $4.65\%$ vs. $5.30\%$).

Our proposed fine-tuning strategies also outperform the previous method in \cite{sun2019fine} on text classification tasks, which makes extensive efforts to find sophisticated hyperparameters.

\paragraph{Effects on Fine-tuning  BERT-Large}\label{experiment-large}

\begin{table}[h!]\small\setlength{\tabcolsep}{2pt}
	\centering
	\begin{tabular}{l |c c c| c c}
	\toprule
	\multirow{2}*{Model} & \multirow{2}*{IMDb} & AG's & \multirow{2}*{Avg. $\Delta$} & \multirow{2}*{SNLI} & \multirow{2}*{$\Delta$} \\
	~ & ~ & News & ~ & ~ & ~ \\
	\midrule
	MT-DNN \cite{liu2019multi} & / & / & / & 91.6 & / \\
	$\mathrm{BERT\text{-}L}$ & \multirow{2}*{4.98} & \multirow{2}*{5.45} & \multirow{2}*{-} & \multirow{2}*{90.9} & \multirow{2}*{-} \\
	(our implementation) & ~ & ~ & ~ \\
	\midrule
	$\mathrm{BERT\text{-}L}_{\mathrm{SDA}} (K=1)$ & 4.66 & 5.21 & 5.62\% & \textbf{91.5} & \textbf{6.59\%} \\
	$\mathrm{BERT\text{-}L}_{\mathrm{SDA}} (K=T-1)$ & \textbf{4.58} & \textbf{5.15} & \textbf{7.02\%} & 91.4 & 5.49\% \\
	\bottomrule
\end{tabular}
\caption{Effects on fine-tuning the BERT-large model ($\mathrm{BERT\text{-}L}$). For IMDb and AG's News, we report test error rate (\%). For SNLI, we report accuracy (\%).
MT-DNN fine-tunes BERT with multi-task learning.}
\label{table-large}
\end{table}

We also investigate whether self-distillation has similar findings for the BERT-large model ($\mathrm{BERT\text{-}L}$), which contains 24 Transformer layers. Due to the limitation of our devices, we only conduct an experiment on two text classification datasets and one NLI datasets and evaluate strategy $\mathrm{BERT}_{\mathrm{SDA}}$, namely self-distillation with averaged BERT as a teacher. We set two different teacher sizes for comparison. As shown in Table \ref{table-large}, self-distillation also gets a significant gain while fine-tuning the BERT-large model. On two text classification tasks, $\mathrm{BERT\text{-}L}_{\mathrm{SDA}} (K=T-1)$  gives better results and the average improvement is $7.02\%$. For NLI task, $\mathrm{BERT\text{-}L}_{\mathrm{SDA}} (K=1)$ gives better result and the improvement is $6.59\%$.

Moreover, although our self-distillation fine-tuning strategy does not leverage the external data or knowledge, it also gives a comparable performance of MT-DNN \cite{liu2019multi}, which fine-tunes BERT with a specific projection layer under the multi-task learning framework.

\paragraph{Discussion}

In general, $\mathrm{BERT}_{\mathrm{SDA}}$ has a similar phenomenon compared to $\mathrm{BERT}_{\mathrm{SDV}}$, while having better
computational and memory efficiency. 
Considering that $\mathrm{BERT\text{-}L}_{\mathrm{SDA}} (K=1)$ is same as $\mathrm{BERT\text{-}L}_{\mathrm{SDV}} (K=1)$, and it performs better than $\mathrm{BERT\text{-}L}_{\mathrm{SDA}} (K=T-1)$. The SDA models are generally worse than SDV models on NLI tasks.
All the above illustrates that parameter averaging is worse than logits voting when dealing with difficult tasks such as NLI, but better on
simple tasks such as text classification.

\section{Conclusion}

In this paper, we propose simple but effective fine-tuning strategies for BERT without external knowledge or data. Specifically, we introduce two mechanisms: self-ensemble and self-distillation. The self-ensemble method with parameter averaging can improve BERT without significantly decreasing the training efficiency. With self-distillation, the student and teacher models can benefit from each other.
Our proposed strategies are orthogonal to the approaches with external data and knowledge. Therefore, we believe that our strategies can be further boosted by more sophisticated hyperparameters and data augmentation.

In future, we will investigate a better fine-tuning strategy by integrating our proposed method into an optimization algorithm.

\bibliographystyle{named}
\bibliography{ijcai20}

\end{document}